\documentclass{article}
\usepackage[english]{babel}
\usepackage[utf8]{inputenc}
\usepackage{johd}
\usepackage{graphicx}
\graphicspath{ {./images/} }
\usepackage{amsmath}
\usepackage{amssymb}
\usepackage{hyperref}
\hypersetup{
    colorlinks=true,
    linkcolor=blue,
    filecolor=magenta,      
    urlcolor=cyan,
    pdfpagemode=FullScreen,
    }
\title{New Designed Loss Functions to Solve Ordinary Differential Equations with Artificial Neural Network}

\author{Xiao Xiong\\
        Imperial College London\\
        \tt{xx1119@ic.ac.uk}
}
\date{}

\begin{document}
\maketitle
\begin{abstract} 
\noindent This paper investigates the use of artificial neural networks (ANNs) to solve differential equations (DEs) and the construction of the loss function which meets both differential equation and its initial/boundary condition of a certain DE. In section 2, the loss function is generalized to $n^\text{th}$ order ordinary differential equation(ODE). Other methods of construction are examined in Section 3 and applied to three different models to assess their effectiveness.
\end{abstract}

\noindent\\
\section{Introduction}
Differential equations are used in the modeling of various phenomena in academic fields. Most differential equations do not have analytical solutions and are instead solved using domain-discretization methods\cite{1}\cite{2}\cite{3} such as boundary-element, finite-differences, or finite-volumes to obtain approximated solutions. However, discretization of the domain into mesh points is only practical for low-dimensional differential equations on regular domains. Furthermore, approximate solutions at points other than mesh points must be obtained through additional techniques such as interpolation.\\

\noindent
Monte Carlo methods and radial basis functions\cite{4} have also been proposed as alternatives for solving differential equations without the need for mesh discretization. These methods allow for the easy generation of collocation points within the domain, but they are not as stable or efficient as mesh-based methods.\\

\noindent
In this article, we introduce the use of artificial neural networks (ANNs) as an alternative method for solving differential equations. This approach does not require complex meshing and can be used as a universal function approximator\cite{5} to produce a continuous and differentiable solution over the entire domain.\\

\noindent
To obtain an exact solution for a differential equation, both the main equation and the constraint equations (initial/boundary conditions) must be taken into account. The key challenge is figuring out how to use one single network to satisfy these equations at the same time. One method is the DGM algorithm\cite{6} shown in Figure 1, which minimizes the direct sum of three individual losses from the main differential equation, the initial conditions, and the boundary conditions in a single neural network. However, the solution obtained 
through this algorithm is not an accurate approximation to the solution of the main and constraint equations, as summing the losses can affect the accuracy of each other.\\
\begin{figure}[h]
\centering
\includegraphics[width=0.5\textwidth]{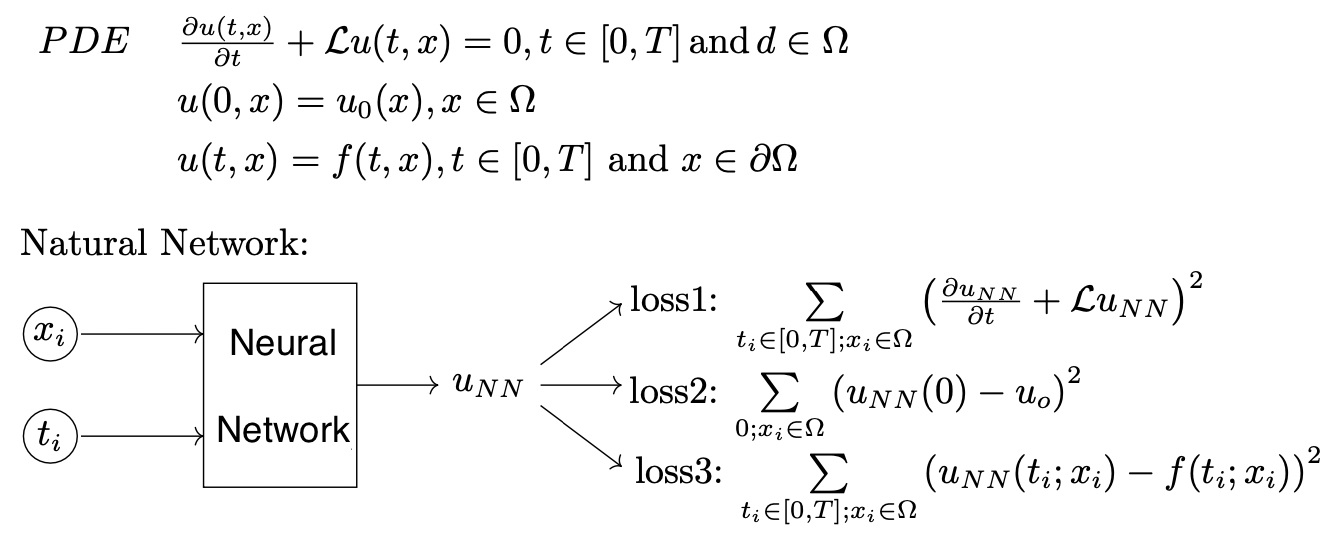}
\caption{DGM algorithm for solving PDE}
\end{figure}

\noindent
Lagaris\cite{7} proposed a method in which the loss of the neural network can be reconstructed to obtain an exact solution to both the differential equation and its constraint conditions.  The details of this method will be introduced and modified in the next section where the general formula of the loss functions for ordinary differential equations (ODEs) of any order is also established. In Section 3, different forms of loss functions are designed and applied to three practical models with related code in \href{https://github.com/XIAOSHERRY/A-New-Designed-Loss-Function-to-Solve-Ordinary-Differential-Equations-with-Artificial-Neural-Network}{GitHub}. Some possible future works based on this paper are provided in the final session.
\section{Construction of Loss Function }
This section explains the concept of differential equations and the process of solving them using multi-layer perceptrons. It also describes how to construct loss functions for these equations during the training of neural networks, including examples for first- and second-order ODEs. Finally, the section discusses how the formula for the loss function can be generalized to nth-order ODEs.\\

\noindent
Firstly, the definition of a general differential equation is 
\begin{equation}
    F(\textbf{x}, u, Du, D^{2}u,...,D^{m}u) = 0, \quad \textbf{x} \in \Omega \subset \mathbb{R}^{n}
\end{equation}
where $u$ is the unknown solution to be determined and 
\begin{equation*}
    D^{m}u = \Bigg\{ 
    \frac{\partial^{|\alpha|} u}{\partial x_{1}^{\alpha_{1}} \dots \partial x_{n}^{\alpha_{n}}}
     \bigg| \alpha \in \mathbb{N}^{n} \text{,}
    |\alpha|=m \bigg|\Bigg\}
\end{equation*}
The inputs to the neural network for solving the differential equation are discretized points of the domain $\Omega$. Then the differential equation becomes a system of equations $F(\textbf{x}_{i}, u(\textbf{x}_{i}), Du, D^{2}u,...,D^{m}u) = 0$ for all $\mathbf{x}_{i} \in \hat{\Omega}$. When a specific neural network is trained, $u_{NN}(\mathbf{x},\mathbf{p)}$ is used to present the output where $\mathbf{p}$ are weights and biases and $\mathbf{x}$ are inputs. The general loss used to do gradient descent is $\min_{\mathbf{p}}\sum_{\mathbf{x}_{i} \in \hat{\Omega} }(F(\textbf{x}_{i}, u_{NN}(\textbf{x}_{i}), Du_{NN}, D^{2}u_{NN},...,D^{m}u_{NN}))^{2}$, subject to initial and boundary conditions.\\

\noindent
In Lagaris's approach\cite{7} mentioned in section 1, he constructed the solution $u_{NN}(\mathbf{x}, \mathbf{p})$ in the loss function to satisfy the differential equation and its initial or boundary conditions as 
\begin{equation}
    u_{NN}(\mathbf{x}, \mathbf{p}) = A(\mathbf{x}) + G(\mathbf{x}, N(\mathbf{x}, \mathbf{p}))
\end{equation}
where N is an output of a feed-forward neural network, term $A(\mathbf{x})$ corresponds to constraint conditions and term G satisfies the differential equation. After implementing the proposed method, solving certain differential equations becomes an unconstrained problem that only involves a single equation, which means that we only need to calculate a single loss in the neural network compared with DGM algorithm to obtain the solution. Note that the gradient computation in this neural network process involves derivatives of the output with respect to any of its inputs (which is used in the calculation of losses) and its parameters (which is used in gradient descent). The procedure of gradient computation is deduced in paper\cite{7}. \\

\noindent
\subsection*{First order ODE}
To specialize equation (2) in the first order ODE
\begin{equation}
    \frac{du}{dt} = f(t,u)
\end{equation}
with an initial condition $u(a)=A$, we let 
\begin{equation}
    u_{NN}(t)=A+(t-a)N(t,\mathbf{p})
\end{equation}
in which $u_{NN}(a) = A$, thus $u_{NN}$ satisfies the initial condition. After generating n inputs $t_{i}$ in the domain, we get n corresponding outputs by training a feed-forward neural network with the same weights and bias for each input value in a single epoch. Then the loss we need to minimize for gradient descent is $L[\mathbf{p}]=\sum_{i} \biggl\{ \frac{du_{NN}(t_{i})}{dt}-f(t_{i}, u_{NN}(t_{i})) \biggr\}^{2}$.

\subsection*{Second order ODE}
There are three loss construction cases corresponding to three different constraint condition cases in second-order ODE:
\begin{equation}
    \frac{d^{2}u}{dx^{2}} = f(x,u,\frac{du}{dx} )
\end{equation}
The constraints in the first case are $ u(a)=A \quad \text{and} \quad \frac{du}{dx}\bigg|_{x=b}=B$ which results in the constructed function
\begin{equation}
    u_{NN}(x) = A + B(x-a) + (x-a)(x-b)^{2}N(x,\mathbf{p})
\end{equation}
$u_{NN}'(b)=B$ can be verified by differentiating the above function with respect to t, getting $u_{NN}'(x) = B + [(x-b)^{2}+2(x-b)(x-a)]N(x,\mathbf{p}) + (x-a)(x-b)^{2}N'$.\\

\noindent
The second case involves two conditions $u(a)=A$ and $u(b)=B$, while the third case contains conditions on $\frac{du}{dx}$, namely $\frac{du}{dx}\bigg|_{x=a}=A$ and $\quad \frac{du}{dx}\bigg|_{x=b}=B$ with corresponding parts of loss function
\begin{equation}
    u_{NN}(x) = A \Bigl( \frac{x-b}{a-b}\Bigr) + B\Bigl(\frac{x-a}{b-a}\Bigr) + (x-a)(x-b)N(x,\mathbf{p})
\end{equation}
and
\begin{equation}
    u_{NN}(x) = \frac{A(x-b)^{2}}{2(a-b)}+\frac{B(x-a)^{2}}{2(b-a)}+(x-a)^{2}(x-b)^{2}N(x,\mathbf{p}).
\end{equation}
Finally, the loss function derived from these constructed $u_{NN}$ function for second-order ODE is similar to the first-order one.

\subsubsection*{$n^{\text{th}}$ order ODE}
After clarifying how to construct functions in $1^{\text{st}}$ and $2^{\text{nd}}$ order cases, I extend formula to $n^{\text{th}}$ order ODE. The general $n^{\text{th}}$ order ODE is written as
\begin{equation}
    \frac{d^{n}u}{dx^{n}} = f(x,u,\frac{du}{dx},...,\frac{d^{n-1}u}{dx^{n-1}})
\end{equation}
with different constraint conditions listed as follows:
\subsubsection*{Case 1}
In this case, conditions on $u$ are considered, equivalently
\begin{equation}
    u(x_{i})=C_{i} \quad \text{for} \quad \{i \in \mathbb{Z}|1 \leq i \leq n\} 
\end{equation}
where $x_{i} \neq x_{j}$ if $i \neq j$.
We reconstruct the solution function as 
\begin{equation}
    u_{NN}(x)=\sum_{i=1}^{n}(C_{i}\prod_{j\neq i}^{j\in{1,2,3,...,n}}\frac{x-x_{j}}{x_{i}-x_{j}}) + \prod_{i=1}^{n}(x-x_{i})N(x,\mathbf{p})
\end{equation}

\subsubsection*{Case 2}
The conditions on $n-1^{\text{th}}$ order derivatives are given in the forms
\begin{equation}
    \frac{d^{n-1}u}{dx^{n-1}} \bigg|_{x=x_{i}}=C_{i} \quad \text{for} \quad \{i \in \mathbb{Z}|1 \leq i \leq n\}
\end{equation}
which leads to the constructed solution
\begin{equation}
    u_{NN}(x)=\sum_{i=1}^{n}C_{i}\frac{M_{i}(x)}{\frac{d^{n-1}M_{i}}{dx^{n-1}}|_{x=x_{i}} } +\prod_{i=1}^{n}(x-x_{i})^{n}N(x,\mathbf{p}) 
\end{equation}
where $M_i(x)=\prod_{j\neq i}^{j\in {1,2,...,n}}(x-x_{j})^{n}$.

\subsubsection*{Case 3}
Case 3 involves conditions on derivatives up to order $n-1$:
\begin{equation}
    \frac{d^{i}u}{dx^{i}}\bigg|_{x=x_{i}}=C_{i} \quad \text{for} \quad \{i \in \mathbb{Z}|0 \leq i \leq n-1\}
\end{equation}
We first define a series of functions $M_{i}(x) = M_{i-1}(x-x_{i-1})^{i}$ with $M_{0}(x)  = 1$ and some coefficients $N_{i}  = \biggl( C_{i}-\sum_{j=o}^{i-1}(M_{j}^{(i)}(x)|_{x=x_{i}}\times  N_{j})\biggl)\times 
\frac{1}{M_{i}^{(i)}|_{x=x_{i}}} $ with $N_{0} = C_{0}$. Then the reconstructed part of the loss function becomes
\begin{equation}
    u_{NN} = \sum_{i=0}^{n-1} N_{i}\times M_{i}(x) + M_{n}N(x, \mathbf{p})
\end{equation}

\subsubsection*{Case 4: General Condition Case}
In this part, I will introduce the general formula for the general condition cases. The format of the condition is 
\begin{equation}
    \frac{d^{i}u}{dx^{i}}\bigg|_{x=x_{i \alpha}}=C_{i \alpha} \quad \text{with} \quad \{i \in \mathbb{Z}|0 \leq i \leq n-1\} \quad \text{and} \quad \{\alpha \in \mathbb{Z}|1 \leq \alpha \leq \alpha_{i}\}
\end{equation}
where $\sum_{i=0}^{n-1}\alpha_{i} = n$. The differential order of conditions in this scenario can vary and the same order differentiated condition can include multiple cases. However, the total number of conditions must be equal to n. Similar to case 3, functions and coefficients are defined in advance by
\begin{equation*}
    M_{i\alpha}(x)=(x-x_{i\alpha})^{i+1}, \quad
    M_{i}(x) = \prod_{\alpha=1}^{\alpha_{i}}M_{i\alpha}, \quad
    F_{i\alpha}(x)=\frac{1}{M_{i\alpha}} \prod_{j=0}^{i}M_{j}
\end{equation*}
and 
\begin{equation*}
     N_{0\alpha} = C_{0\alpha}\times\frac{1}{F_{0\alpha}(x)|_{x=x_{0\alpha}}}, \quad
     N_{i\alpha}  = \biggl(C_{i\alpha}-\sum_{j=0}^{i-1}\sum_{\beta=1}^{\alpha_{j}}N_{j\beta}F_{j\beta}^{(i)}|_{x=x_{i\alpha}}-\sum_{\beta\neq\alpha}^{\beta\in{1,...,\alpha_{i}}}N_{i\beta}F_{i\beta}^{(i)}|_{x=x_{i\alpha}}\biggl)\times
\frac{1}{F_{i\alpha}^{(i)}(x)|_{x=x_{i\alpha}}}
\end{equation*}
Finally, the general reconstructed solution function formula for $n^{\text{th}}$ order ODE with general constraint conditions is 
\begin{equation}
    u_{NN} = \sum_{i=0}^{n-1}\sum_{\alpha=1}^{\alpha_{i}}N_{i\alpha}\times F_{i\alpha}(x)
+ \prod_{i=0}^{n-1}M_{i}(x)N(x,\mathbf{p}) 
\end{equation}

\subsubsection*{System of K ODEs}
Now, the system of ODE involving K equations is explored. We start from the first-order system ODE
\begin{equation}
    \frac{du_{k}}{dx}=f_{k}(x,u_{1},...,u_{K}) \quad \text{for}\quad k \in {1,2,...,K}
\end{equation}
together with conditions $u_{k}(x_{k})=C_{k}$. In the ANN approach for solving the ODE system, total K multi-layer perceptrons work in parallel to process K equations. For each first-order equation, we get a reconstructed function

\begin{equation}
    u_{NN_{k}}(x)=A_{k}+xN_{k}(x,\mathbf{p}_{k})
\end{equation}
and the loss for each neural network is $L[\mathbf{p}_{k}]=\sum_{i}\bigg(\frac{du_{NN_{k}}}{dx}\bigg|_{x=x_{i}}-f_{k}(x_{i},u_{NN_{1}},u_{NN_{2}},...,u_{NN_{K}})\bigg)^{2}$.\\

\noindent
To generalization, the system of K nth-order ODEs is shown as \begin{equation}
    \frac{d^{n}u_{k}}{dx^{n}}=f_{k}(x,S_{0},...,S_{n-1}) \quad \text{for} \quad  k \in {1,2,...,K}
\end{equation}
 with the notation $ S_{i}={\frac{d^{i}u_{1}}{dx^{i}},\frac{d^{i}u_{2}}{dx^{i}},...,\frac{d^{i}u_{K}}{dx^{i}}} $ for $\quad i \in 0,...,n-1$. The constraint conditions are $\frac{d^{i}u_{k}}{dx^{i}}\bigg|_{x=x_{ki\alpha}}=C_{ki\alpha}$ where $i$'s are integers in the range $[0 \leq i \leq n-1]$ and $\alpha$'s are integers in the range $[1 \leq \alpha \leq \alpha_{ki}]$ with $\sum_{i=0}^{n-1}\alpha_{ki} = n$. The reconstructed function for each equation in the system is exactly the same as $n^{\text{th}}$ order single ODE in equation(17).

\section{Application in models}
In this section, we will look at different methods for reconstructing $u_{NN}$ in loss functions for first-order ordinary differential equations. Based on these methods, We can reconstruct solutions to higher-order ODEs, similar to how we did in section 2. There are a total of seven different constructions shown in Table 1. The polynomial construction is an extension of the one presented in equation(4), where an additional coefficient c is included and adjusted based on the specific ODE being considered. The exponential construction was introduced by Chen in \cite{8}. Further to the above two constructions, I propose five more forms. The impact of the base of logarithm will also be examined in the next part of the analysis.\\ 

\begin{table}[!h]
\begin{center}
\begin{tabular}{||c|c||} 
 \hline
 Function name & Formula \\ [0.5ex] 
 \hline\hline
 Polynomial & $u_{NN}(x)=A+c(x-a)N(x,\mathbf{p})$\\ 
 \hline
 Exponential & $u_{NN}(x)=A+(1-e^{-(x-a)})N(x,\mathbf{p})$\\
 \hline
 Hyperbolic & $u_{NN}(x)=A+\frac{e^{(t-a)}-e^{(a-t)}}{e^{(t-a)}+e^{(a-t)}}N(x,\mathbf{p})$\\
 \hline
 Logarithmic & $u_{NN}(x)=A+log_{c}(t+1-a)N(x,\mathbf{p})$ \\
 \hline
 Logistic & $u_{NN}(x)=A+(\frac{1}{1+e^{(-t+a)}}-\frac{1}{2})N(x,\mathbf{p})$ \\
 \hline
 Sigmoid & $u_{NN}(x)=A+(\frac{t-a}{1+e^{(a-t)}})N(x,\mathbf{p})$ \\
 \hline
 Softplus & $u_{NN}(x)=A+(ln(1+e^{(t-a)})-ln(2))N(x,\mathbf{p})$\\ [1ex] 
 \hline
\end{tabular}
\caption{\label{demo-table}7 forms of solution reconstruction}
\end{center}
\end{table}

\noindent
For second-order ODE with initial condition $u(a)=A \quad \text{and} \quad \frac{du}{dx}\bigg|_{x=b}=B$, the exponential construction is 
\begin{equation*}
    u_{NN}(t)=A+B(t-b)+(1-e^{-t-a)})^{2}N(x,\mathbf{p})
\end{equation*}
In the following model analysis, the performance of the seven functions mentioned above will be evaluated when they are implemented using a neural network approach in three different models. The average of the 100 lowest losses within a certain number of epochs will be calculated to evaluate their performance. Additionally, the trends of loss value decreasing during epochs will be plotted and compared. Finally, the solution of the reconstructed function with the lowest loss will be compared with the analytical solution to state the accuracy of the ANN approach.

\subsection{Newton's Law of Cooling}
\subsubsection*{Model Description}
Newton's law of cooling\cite{9} states that the rate of change of the temperature of an object is proportional to the difference between its own temperature and the temperature of its surroundings. This relationship is described by the equation:
\begin{equation}
    \frac{dT}{dt} = r (T_{env} - T(t))
\end{equation}
where
\begin{itemize}
    \item $\frac{dT}{dt}$ is the rate of change of temperature with respect to time (also known as the cooling rate)
    \item $T$ is the temperature of an object
    \item $T_{env}$ is the temperature of the environment
    \item $r$ is the cooling coefficient, which is a constant that depends on the characteristics of the object and the environment it is in
\end{itemize}
In this study, we will predict how the temperature of the boiling water will change over time, given an initial temperature of $100^\circ C$ and a cooling coefficient of 0.5, in an environment with a temperature of $10^\circ C$.\\

\noindent
The analytical solution to the initial-value problem described above, solved using separation of variables, is as follows:
\begin{equation}
    T(t) = 10 + 90e^{-0.5t}
\end{equation}

\subsubsection*{Evaluation of Function Performance}
In this model, the specific neural network was trained for 200000 epochs in order to obtain the average losses shown in Table 2.
\begin{table}[!h]
\begin{center}
\begin{tabular}{||c|c| c| c| c| c| c |c||} 
 \hline
     & polynomial & exponential & hyperbolic & logarithm & logistic & sigmoid & softplus\\ [0.5ex] 
 \hline\hline
 Average Loss & 2.19e-05 & 1.55e-07 & 1.03e-06 &1.75e-06&6.67e-07&1.84e-05&1.78e-05\\ 
 \hline
\end{tabular}
\caption{\label{demo-table}Average losses of 7 constructed functions in the model of  Newton's law of cooling }
\end{center}
\end{table}
The loss trends are also plotted in Figure 2.
\begin{figure}[h]
    \begin{minipage}{0.5\textwidth}
        \centering
        \includegraphics[width=\linewidth]{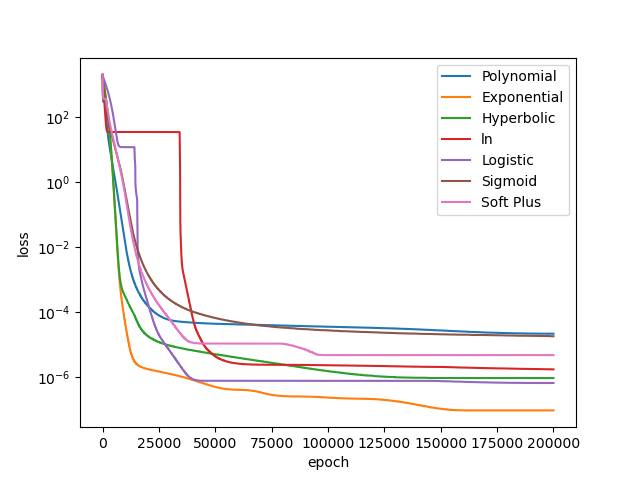}
        \caption{Comparing losses of 7 constructed functions in model of Newton's law of cooling}
    \end{minipage}%
    \begin{minipage}{0.5\textwidth}
        \centering
        \includegraphics[width=\linewidth]{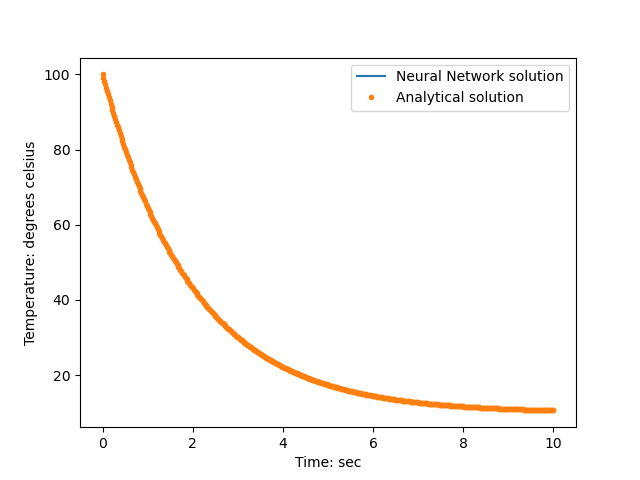}
        \caption{Comparing ANN and analytical solutions in model of Newton's law of cooling}
    \end{minipage}
\end{figure}
Based on the data presented in the table and graph, it can be seen that the exponential function performs the best in this model due to its ability to quickly reach the lower loss value and to achieve the lowest at the end of epochs. The logistic function also performs well in this model and reaches its own lowest loss value at an early point in the epochs.\\

\noindent
Finally, we compare our ANN solution based on the exponential function and the analytical solution in Figure 3, which demonstrates the effectiveness of the ANN approach.

\subsection{Motor Suspension System}
\subsubsection*{Model Description}
A motor suspension system\cite{10} is a mechanical system that is used to support and isolate the motorcycle wheels from the rest of a vehicle. A mass-spring-damper model can be used to model the behavior of a motor suspension system by representing the wheels of a motorcycle as a mass, the suspension system as a spring, any damping effects as a damper, and the shock absorber attached to the suspension system as a fixed place. When the motorcycle wheels are resting on the ground and the rider gives extra force to them by sitting on the motor, the spring becomes compressed, bringing the system into an equilibrium position. What we are investigating now is the motion (displacement from the equilibrium position)of the wheels after the motor has a jump.\\

\noindent
The modeling differential equation is 
\begin{equation}
    m\frac{dx^2}{dt}+c\frac{dx}{dt}+kx=0
\end{equation}
where
\begin{itemize}
    \item $m$ is the mass of the motorcycle wheels
    \item $c$ is the damping coefficient
    \item $k$ is the spring constant
    \item $x$ is the displacement of the motor from its equilibrium position
    \item $\frac{dx}{dt}$ is the velocity of the motor when hitting the ground
    \item $\frac{dx^2}{dt}$ is the acceleration of the motor
\end{itemize}
In the English system, the total mass of wheels and a rider is chosen to be $m = 12$ slogs, the spring constant is $k = 1152$, the damping constant is $c = 240$, the initial displacement is $x_0 = \frac{1}{3}$ft and the initial velocity is $\frac{dx_0}{dt}$ = 10ft/sec.\\

\noindent
Solving the above differential equation using a characteristic equation gives an analytical solution
\begin{equation}
    x(t) = 3.5e^{-8t}-\frac{19}{6}e^{-12t}
\end{equation}

\subsubsection*{Evaluation of Function Performance}
In this model, the neural network is run for 200000 epochs, giving the average lowest losses for seven reconstructed functions in Table 3.\\
\begin{table}[!h]
\begin{center}
\begin{tabular}{||c|c| c| c| c| c| c |c||} 
 \hline
     & polynomial & exponential & hyperbolic & logarithm & logistic & sigmoid & softplus\\ [0.5ex] 
 \hline\hline
 Average Loss & 3.26e-1 & 7.98e3 & 1.32e3 & 2.07e-3 & 4.17e4 & 1.09e2 &
 1.20e5\\
 \hline
\end{tabular}
\caption{\label{demo-table}Average losses of 7 constructed functions in the model of motor suspension system}
\end{center}
\end{table}

\noindent
According to the table, logarithmic and polynomial functions have lower losses compared to the other functions shown. This is also evident in Figure 4.\\
\begin{figure}[h]
    \begin{minipage}{0.5\textwidth}
        \centering
        \includegraphics[width=\linewidth]{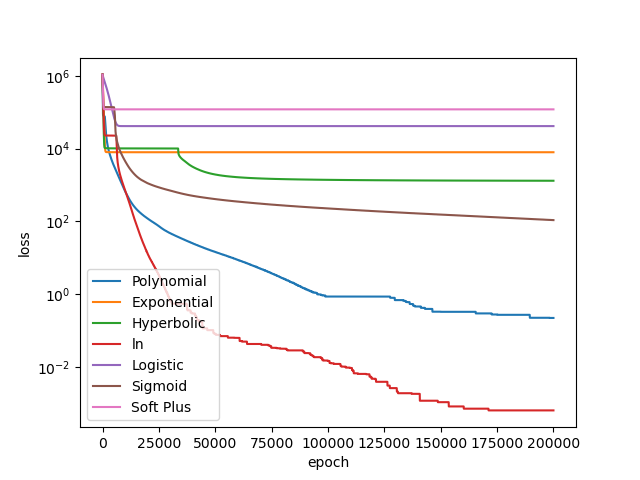}
        \caption{Comparing losses of 7 constructed functions in model of motor suspension system}
    \end{minipage}%
    \begin{minipage}{0.5\textwidth}
        \centering
        \includegraphics[width=\linewidth]{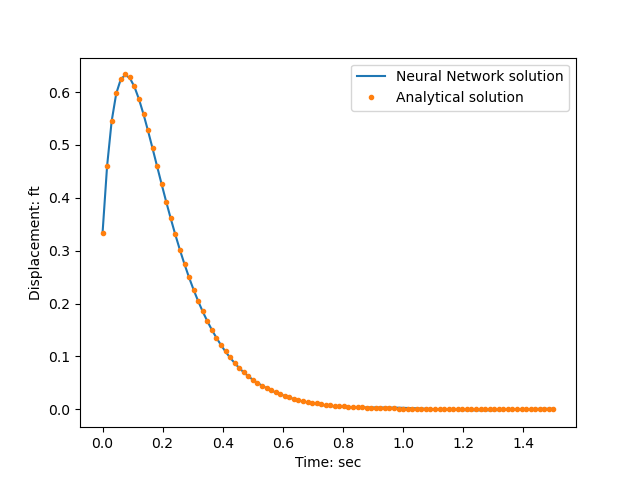}
        \caption{Comparing ANN and analytical solutions in model of motor suspension system}
    \end{minipage}
\end{figure}

\noindent
Figure 5 demonstrates that the numerical solution produced by our artificial neural network aligns perfectly with the analytical solution in this model.

\subsubsection*{Evaluation of polynomial functions with different coefficients}
In this section, we will investigate how the coefficient of the polynomial function impacts the loss performance of a neural network in this  motor suspension system. 
\begin{figure}[h]
\centering
\includegraphics[width=0.5\textwidth]{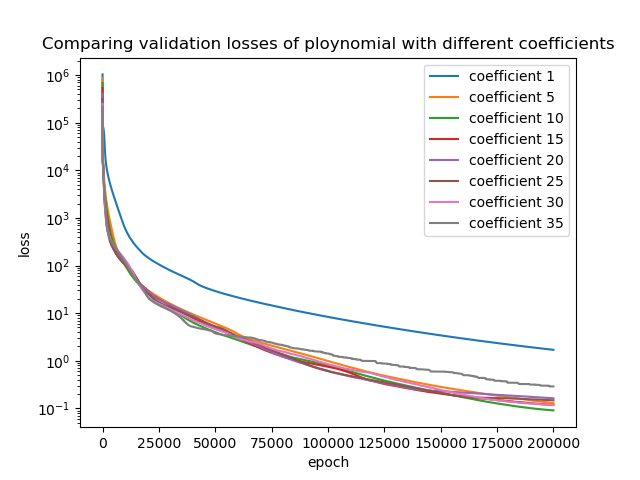}
\caption{Comparing losses of polynomial functions with different coefficients}
\end{figure}
Notice in Figure 6, when the coefficients are very small or large, the loss that is achieved after completing all epochs tends to be higher. The impact of coefficient values between 5 and 30 is roughly the same. In other words, we can select any coefficient within the above range in order to obtain an accurate solution.

\subsection{Home Heating}
\subsubsection*{Model Description}
The heat transfer in a house can be modeled using a system of first-order ordinary differential equations\cite{11}. This is useful for understanding how the temperature of a house changes over time, and for designing heating systems that maintain a comfortable temperature inside the house.\\

\noindent
Here, we examine the variations in temperature of the attic, basement, and insulated main floor in a house using Newton's cooling law. It is assumed that the temperature outside, in the attic, and on the main floor is constantly $35^\circ F$ during the day in the winter, and the temperature in the basement is $45^\circ F$ before the heater is turned on. When a heater starts to work at noon (t=0) and is set the temperature to $100^\circ F$, it increases the temperature by $20^\circ F$ per hour. This could be modeled as following differential equations:
\begin{align}
    \frac{dx}{dt} & = k_0(T_{earth}-x)+k_1(y-x)\\
    \frac{dy}{dt} & = k_1(x-y)+k_2(T_{out}-x)+k_3(z-y)+Q_{heater}\\
     \frac{dz}{dt} & = k_3(y-z)+k_2(T_{out}-x)
\end{align}
where
\begin{itemize}
    \item $x$, $y$, and $z$ represent the temperatures of the basement, main living area, and attic, respectively
    \item $T_{earth}$,$T_{out}$ represent the initial temperatures of the basement and outside respectively.
    \item The variable $Q_{heater}$ represents the heating rate of the heater 
    \item The k is called the cooling constant which describes the rate of change of the temperatures of each area over time.
\end{itemize}
As the model described above, the initial temperatures at noon (t = 0) are $T_{earth}$ = x(0) = 45, $T_{out}$= y(0) = z(0) = 35 and $Q_{heater}=20$. And we set cooling constant as $k_{0}=\frac{1}{2}$, $k_{1}=\frac{1}{2}$, $k_{2}=\frac{1}{4}$, $k_{3}=\frac{1}{4}$, $k_{4}=\frac{3}{4}$.

\subsubsection*{Evaluation of Function Performance}
This time, we train a certain neural network 20000 times to achieve a low loss value at the end. The average of the 100 lowest losses and how losses decrease during the training process for different constructed functions are displaced in table 4 and Figure 7.
\begin{table}[!h]
\begin{center}
\begin{tabular}{||c|c| c| c| c| c| c |c||} 
 \hline
     & polynomial & exponential & hyperbolic & logarithm & logistic & sigmoid & softplus\\ [0.5ex] 
 \hline\hline
 Average Loss & 3.41e-04 & 1.23e-06 & 1.19e-05 & 4.48e-06 & 2.36e-04 & 2.69e-03 & 2.12e-03\\ 
 \hline
\end{tabular}
\caption{\label{demo-table}Average losses of 7 constructed functions in Home Heating model }
\end{center}
\end{table}

\begin{figure}[h]
    \begin{minipage}{0.5\textwidth}
        \centering
        \includegraphics[width=\linewidth]{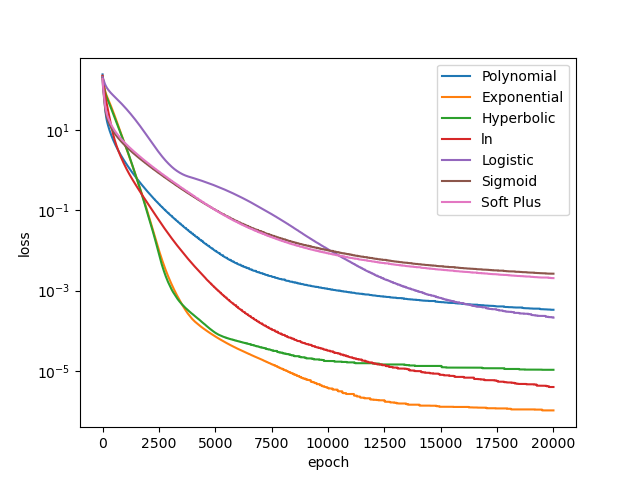}
        \caption{Comparing losses of 7 constructed functions in home heating model}
    \end{minipage}%
    \begin{minipage}{0.5\textwidth}
        \centering
        \includegraphics[width=\linewidth]{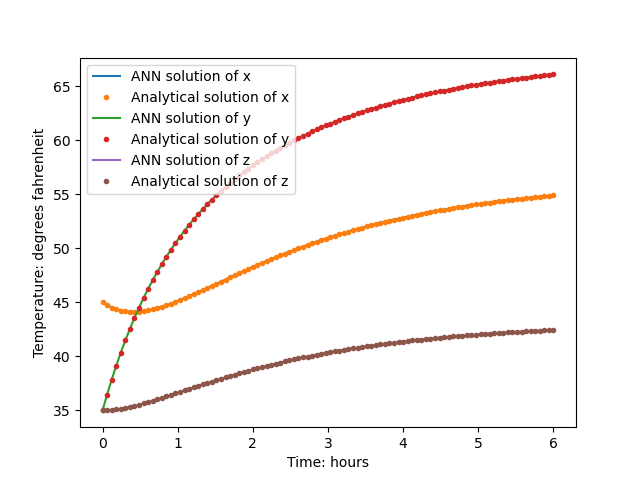}
        \caption{Comparing ANN and analytical solutions in model of home heating}
    \end{minipage}
\end{figure}
\noindent
As shown in the table and graph, the exponential, logarithmic, and hyperbolic functions all have small losses, with the hyperbolic and exponential functions converging the fastest in the first 5000 epochs.\\

\noindent
In Figure 8, the analytical solution of this system of ordinary differential equations is calculated using the "odeint" Python module, which is then compared to the solution obtained using artificial neural networks.

\subsubsection*{Evaluation of logarithmic functions with different basis}
In this section, the effect of the base of logarithm on the loss performance of the ANN approach in the home heating model will be explored. To clearly demonstrate the effect of the base number on the performance, we trained the neural network for 50000 epochs and displayed the results in Figure 9.
\begin{figure}[h]
\centering
\includegraphics[width=0.5\textwidth]{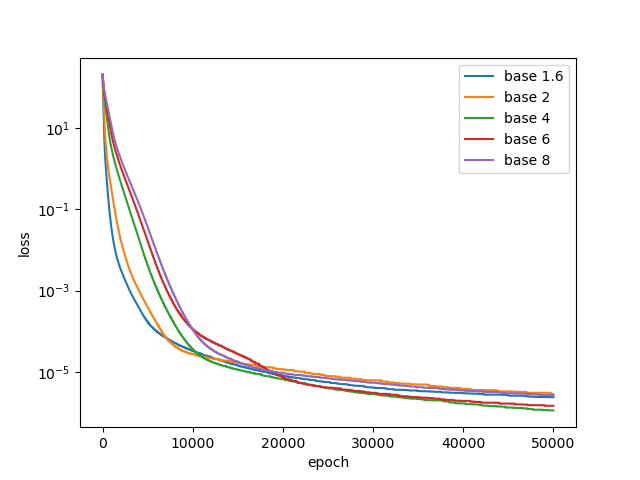}
\caption{Comparing losses of logarithmic functions with different basis}
\end{figure}
It can be observed that the convergence speed is faster when the base number is smaller in the first 10000 epochs. Additionally, logarithm with base 4 results in the lowest loss value in this model, thus being the best choice of base.

\section{Conclusion and Future Research}

Based on the artificial neural network, solving differential equations becomes more accurate and efficient due to the easy generation of domain points and the good approximation capabilities of the neural network. In the paper, the loss function is reconstructed to meet initial/boundary conditions during the artificial neural network process, transforming the constrained problem into an unconstrained one and resulting in a good approximation of the solution. This paper explains how to reconstruct the solution function and extends the construction formula to $n^{\text{th}}$ order ODEs. In addition, the different forms of construction are tested in three realistic models to evaluate their effectiveness.\\

\noindent
The general $n^{\text{th}}$ order formula is only based on the polynomial function for the ordinary differential equations. As future work, the other six forms of reconstruction mentioned in section 3 could be extended to $n^{\text{th}}$ order ODEs. Meanwhile, the development of a general construction formula for partial differential equations could be pursued to better fit a wider range of models in various academic fields.

\end{document}